\documentclass[11pt,a4paper]{article}
\usepackage[hyperref]{acl2021}
\usepackage{times}
\usepackage{latexsym}

\usepackage{microtype}

\aclfinalcopy 
\def\aclpaperid{} 

\usepackage{graphicx}

\usepackage{multirow}

\usepackage{subcaption}

\usepackage{amsmath}
\usepackage{amssymb} 

\usepackage{makecell}

\usepackage{xcolor}

\usepackage{arydshln}

\usepackage{xurl}
\hypersetup{breaklinks=true}
\usepackage[hyphenbreaks]{breakurl}

\title{Can current NLI systems handle German word order? Investigating language model performance on a new German challenge set of minimal pairs}

\author{Ines Reinig \\
  Data and Web Science Group \\
  Mannheim University \\
 Germany\\
  \texttt{ines.reinig@uni-mannheim.de} \\\And
  Katja Markert \\
  Institute of Computational Linguistics \\
  Heidelberg University \\
 Germany\\
  \hspace*{1em}\texttt{markert@cl.uni-heidelberg.de} \\}

\date{}

\begin{document}
\maketitle
\begin{abstract}
Compared to English, German word order is freer and therefore poses additional challenges for
natural language inference (NLI). We create WOGLI (Word Order in German Language Inference), the first adversarial  NLI dataset for German word order  that has the following properties: (i)  each premise has an entailed and a non-entailed hypothesis; (ii) premise and hypotheses differ only in word order
and necessary morphological changes to mark case and number. In particular, each premise and its two hypotheses contain 
exactly the same lemmata. 
Our adversarial examples require the model to use morphological markers in order to recognise or reject entailment. 
We show that current German autoencoding models  fine-tuned on translated NLI data can struggle on this challenge set, reflecting the fact that translated NLI datasets will not mirror all necessary language phenomena in the target language. We also examine performance after data augmentation as well as on related word order phenomena derived from WOGLI. Our datasets are publically available at \url{https://github.com/ireinig/wogli}.
\end{abstract}

\section{Introduction}

German is endowed with a rather free word order \citep{bader2019givenness}, especially when it comes to ordering nominal arguments in a sentence.
Currently, large German NLI datasets are only available as translations from other languages.
For example, the training portion (392k pairs) of the German XNLI dataset \citep{conneau-etal-2018-xnli} is a machine translation of the English MultiNLI training set \citep{williams-etal-2018-broad}. The testing portion of German XNLI is a manual translation of 5k English premise-hypothesis pairs that were newly created by the authors of XNLI.
Such translated sets do not necessarily mirror all German-specific linguistic phenomena, such as the freer German word order.


We construct a new German challenge set named WOGLI (Word Order in German Language Inference).
This dataset is handcrafted and does not stem from translation. It contains 16k premises where  each premise is accompanied by one entailed (E) and one non-entailed (NE) hypothesis that both contain the same lemmata as the premise but
change argument order. Morphological markers are indicative of subject and (direct) object, thus informing about the hypothesis' entailment relationship to the premise. In other words, WOGLI serves as a test bed for current language models' capabilities to distinguish subject from object in the context of German word order. 

Our contributions are as follows: 
\begin{enumerate}
    \item We propose the first NLI dataset that specifically targets  German word order phenomena.
    \item We show that current German autoencoding models fine-tuned on the translated XNLI dataset
     can struggle on our proposed challenge set (Sections \ref{sec:results-wogli} and \ref{sec:error-analysis}), tending to always predict entailment for both hypotheses.

    \item We show that data augmentation can help performance on WOGLI but needs a considerable number of examples to work (Section \ref{sec:data-aug-wogli}).
    
    \item We  derive generalization sets including similar word order phenomena to WOGLI to investigate how the augmented models transfer to these datasets and show that
    German word order remains challenging in NLI
      (Section \ref{sec:generalization-wogli}). 
    
\end{enumerate}

\noindent All our datasets 
are
publically available\footnote{\url{https://github.com/ireinig/wogli}}. 

\begin{table*}[!htb]
    \centering
    \begin{tabular}{ll@{\hskip 0.2in}llll@{\hskip 0.2in}l}
        \hline
        Clause & Order & Prefield & L brack. & Middlefield & R brack. & Count (\% of accus.)  \\
        \hline
        Main & SO & \textbf{Peter} & sieht & den Mann & & 231 (86\%) \\
         & & \textbf{Peter} & sees & the man$_{ACC}$ \\
         & & \textit{Peter} & \textit{sees} & \textit{the man} \\
         \cline{2-7}
         & OS & Den Mann & sieht & \textbf{Peter} & & 38 (14\%) \\
         & & The man$_{ACC}$ & sees & \textbf{Peter} \\
         & & \textit{Peter} & \textit{sees} & \textit{the man} \\
        \hline
        Emb. & SO & & dass & \textbf{Peter} den Mann & sieht & 546 (99\%) \\
         & & & that & \textbf{Peter} the man$_{ACC}$ & sees \\
         & & & \textit{that} & \textit{Peter} \textit{sees} \textit{the man} \\
         \cline{2-7}
         & OS & & dass & den Mann \textbf{Peter} & sieht & 6 (1\%) \\
         & & & that & the man$_{ACC}$ \textbf{Peter} & sees \\
         & & & \textit{that} & \textit{Peter} \textit{sees} \textit{the man} \\
        \hline
    \end{tabular}
    \caption[Examples illustrating word order in German clauses and corpus statistics from the literature]{Examples for word order in declarative, active German main and embedded clauses with subject and (accusative) direct object arguments, with corpus statistics from \citet{bader2010word}. As in the remainder of this paper, the subject is always bold. Transliterations and translations (in italics) are provided below each example.}
    \label{tab:german-word-order}
\end{table*}

\section{German Word Order}\label{sec:background}
\paragraph{The topological model.}
The topological model \citep{drach_1937}  describes regularities
in German word order, dependent on the concepts of \textit{prefield} and \textit{middlefield} for
constituent positioning.
In this model, so-called \textit{left and right brackets} form ``[t]he skeleton of the sentence'' \citep[p.~719]{bader2010word}, while other fields are defined according to the position of the verb \citep{durscheid2012syntax}.

Declarative main clauses, such as \textit{Peter sieht den Mann} at the top of Table~\ref{tab:german-word-order}, have a verb-second order. The left bracket contains the finite verb and the prefield is filled with one constituent \citep{bader2010word,durscheid2012syntax}. 
In contrast,  embedded clauses, such as \textit{dass Peter den Mann sieht} in the bottom half of Table \ref{tab:german-word-order}, have a verb-last order. In verb-last clauses, the left bracket is occupied by a subjunction, the right bracket by a finite verb or a verb complex, and other constituents are placed in the middlefield \citep{durscheid2012syntax}. 

While subject followed by object (SO) is viewed as the canonical word order, it is possible to place the object before the subject (OS) in both embedded and main clauses (Table~\ref{tab:german-word-order}). In the main clause either the subject or object
is placed in the prefield, in embedded clauses both are placed in the middlefield but in varying order.

\paragraph{OS acceptability and minimal pairs.} The marked OS order is  more frequent in main clauses involving the prefield \citep{bader2010word} (around 14\% of main clauses with accusative direct object) and in the active voice \citep{bader2017filling} (see data and  examples in Table~\ref{tab:german-word-order}). Therefore, 
we construct our challenge set using only such clauses to raise acceptability of the marked OS word order examples. 
Even in the prefield, OS order can vary in acceptability dependent on relative constituent weight \citep{siewierska1993} (shorter before longer), discourse properties such as givenness \citep{bader2019givenness} (given before new) and semantic properties such as agency \citep{siewierska1993,bader2010word} (animate before inanimate). As we focus on simple grammatical examples without further interference, however, all our constituents are short and  all premises and hypotheses are single sentences. To ensure that entailed and non-entailed sentences are semantically plausible, all our constituents refer to persons. 


\paragraph{German word order in XNLI.}

We extract hypotheses in the training portion of the translated German XNLI (henceforth, GXNLI-train) that are declarative main clauses with a length between 4 and 9 tokens. 
The 38,090 extracted clauses are in active voice and contain one subject NP and one direct object NP in accusative case. We exclude clauses that start with prepositions or adverbs to limit ourselves
to prefield cases. 
Only 1.8\% (698 clauses) of the extracted clauses are in OS order, compared to the 14\% to be expected in a German corpus according to \citet{bader2010word}.
Additionally, a vast majority of the 698 OS clauses start with the same demonstrative pronoun object \textit{das}/\textit{this}, e.g. \textit{Das werde ich tun}/\textit{This I will do}, thus offering little variety.
The extreme prevalence of the SO order in GXNLI-train hypotheses may be due to its translated nature. 

\section{WOGLI  construction}\label{sec:construction-wogli}

\paragraph{Verb Collection.} We collected 50 frequent German transitive verb types including agentive (such as \textit{warnen/warn}), object-experiencer (such as \textit{erschrecken/startle}) and subject-experiencer (such as \textit{lieben/love}) verbs. All verbs  can take animate (human) subjects as well as animate
(human) direct objects, and all objects
take the accusative case. All verbs are not symmetric, meaning that they do not lead to bidirectional entailments.\footnote{For example, for the symmetric verb \textit{heiraten}/\textit{marry}, X marries Y would entail Y marries X, which would not allow us to automatically derive non-entailed hypotheses.}
In addition, none of the verbs  need to split prefixes when used in  main clauses so that the resulting premises have a very simple SVO structure. 
All verbs occur at least 70 times in GXNLI-train. Consequently, any difficulties that a language model will experience are unlikely to be due to verb rarity.

\paragraph{Noun Collection.} We collected 144 noun types  describing humans that  function as direct object or subject in our premises/hypotheses.  These include 38 masculine common nouns such as \textit{Gast/guest},
each of which was seen at least 10 times
in GXNLI-train and
24 feminine common nouns such as \textit{Lehrerin/(female) teacher}. We collected  feminine common nouns by searching for the suffix \textit{in} in GXNLI-train, which often indicates female persons in German. The unbalanced masculine-feminine split is due to the automatic translation of GXNLI-train as gender-neutral English job descriptions, for example \textit{doctor}, are most frequently  translated via the German male form, e.g. \textit{Arzt} instead of the female form \textit{\"Arztin}\footnote{We could have made up the shortfall by including more feminine forms, even if they do not occur in 
GXNLI-train, but we consider it more important for this study to keep lexical differences to the fine-tuning set minimal.}. We also collected 41 female and 41 male first names
that occur at least 10 times in GXNLI-train. The 144 noun types yield 181 different noun surface forms (nominative/accusative, plural/singular).


\paragraph{Premise and Hypothesis Generation.} We automatically generated German  premises as declarative, present tense, main clauses in the active voice with  
SVO structure (see lines 1 and 5) in Table~\ref{tab:wogli-examples}).  
Each SVO premise is accompanied by two hypotheses. H1-SO (NE)  exchanges object and  subject including changing S/O case markers and potentially verb number markers. Therefore, similarly to English, this change leads to non-entailment, as the premise \textit{The doctor warns the client} and 
the corresponding H1
\textit{The client warns the doctor} illustrate. We call this subset WOGLI-SO, as the new subject precedes the object.  H2-OS (E)  simply swaps  argument order but keeps case and number markers intact, leading to a sentence synonymous to the premise but with marked OS word order. The resulting set of entailed hypotheses is called WOGLI-OS.
Table~\ref{tab:wogli-examples} shows two full examples with case and number marking. 

\setlength\dashlinedash{1pt}
\setlength\dashlinegap{1.5pt}
\setlength\arrayrulewidth{0.3pt}

\begin{table*}[!htb]
\resizebox{\textwidth}{!}{
 \begin{tabular}{llllll}
  & \multicolumn{5}{l}{} \\
 \hline
 Premise & \underline{\textbf{Der/Dieser/Ein}}$_{NOM-SG-M}$ & \textbf{Arzt} & warnt$_{SG}$ & \underline{den/diesen/einen}$_{ACC-SG-M}$ & \underline{Kunden}\(^{\dagger}\) \\
  & \textbf{The/This/A}$_{NOM-SG-M}$ & \textbf{doctor} & warns$_{SG}$ & the/this/a$_{ACC-SG-M}$ & client \\
  & \textit{The/This/A} & \textit{doctor} & \textit{warns} & \textit{the/this/a} & \textit{client} \\
 \hline
 H1-SO (NE) & \underline{\textbf{Der/Dieser/Ein}}$_{NOM-SG-M}$ & \underline{\textbf{Kunde}}\(^{\dagger}\) &
 warnt$_{SG}$ & \underline{den/diesen/einen}$_{ACC-SG-M}$ & Arzt \\
  & \textbf{The/This/A}$_{NOM-SG-M}$ & \textbf{client} & warns$_{SG}$ & the/this/a$_{ACC-SG-M}$ & doctor \\
  & \textit{The/This/A} & \textit{client} & \textit{warns} & \textit{the/this/a} & \textit{doctor} \\
  \hdashline
 H2-OS (E)* & \underline{Den/Diesen/Einen}$_{ACC-SG-M}$ & \underline{Kunden}\(^{\dagger}\) & warnt$_{SG}$ & \underline{\textbf{der/dieser/ein}}$_{NOM-SG-M}$ & \textbf{Arzt} \\
  & The/This/A$_{ACC-SG-M}$ & client & warns$_{SG}$ & \textbf{the/this/a}$_{NOM-SG-M}$ & \textbf{doctor} \\
  & \textit{The/This/A} & \textit{doctor} & \textit{warns} & \textit{the/this/a} & \textit{client} \\
 \hdashline
 H3-OS (NE)* & \underline{Den/Diesen/Einen}$_{ACC-SG-M}$ & Arzt & warnt$_{SG}$ & \underline{\textbf{der/dieser/ein}}$_{NOM-SG-M}$ & \underline{\textbf{Kunde}}\(^{\dagger}\)\\
 & The/This/A$_{ACC-SG-M}$ & doctor & warns$_{SG}$ & \textbf{the/this/a}$_{NOM-SG-M}$ & \textbf{client} \\
 & \textit{The/This/A} & \textit{client} & \textit{warns} & \textit{the/this/a} & \textit{doctor} \\
\hline
\hline
 Premise & \underline{\textbf{Der/Dieser/Ein}}$_{NOM-SG-M}$ & \textbf{Minister} & \underline{empfiehlt}$_{SG}$ & die/diese$_{ACC-PL-F}$ & Autorinnen \\
  & \textbf{The/This/A}$_{NOM-SG-M}$ & \textbf{minister} & recommends$_{SG}$ & the/these$_{ACC-PL-F}$ & authors \\
  & \textit{The/This/A} & \textit{minister} & \textit{recommends} & \textit{the/these} & \textit{authors} \\
 \hline
 H1-SO (NE) & \textbf{Die/Diese}$_{NOM-PL-F}$ & \textbf{Autorinnen} & \underline{empfehlen}$_{PL}$ & \underline{den/diesen/einen}$_{ACC-SG-M}$ & Minister \\
  & \textbf{The/These}$_{NOM-PL-F}$ & \textbf{authors} & recommend$_{PL}$ & the/this/a$_{ACC-SG-M}$ & minister \\
  & \textit{The/These} & \textit{authors} & \textit{recommend} & \textit{the/this/a} & \textit{minister} \\
  \hdashline
 H2-OS (E)* & Die/Diese$_{ACC-PL-F}$ & Autorinnen & \underline{empfiehlt}$_{SG}$ & \underline{\textbf{der/dieser/ein}}$_{NOM-SG-M}$ & \textbf{Minister}\\
  & The/These$_{ACC-PL-F}$ & authors & recommends$_{SG}$ & \textbf{the/this/a}$_{NOM-SG-M}$ & \textbf{minister} \\
  & \textit{The/This/A} & \textit{minister} & \textit{recommends} & \textit{the/these} & \textit{authors} \\
\hdashline
 H3-OS (NE)* & \underline{Den/Diesen/Einen}$_{ACC-SG-M}$ & Minister & \underline{empfehlen}$_{PL}$ & \textbf{die/diese}$_{NOM-PL-F}$ & \textbf{Autorinnen} \\
  & The/This/A$_{ACC-SG-M}$ & minister & recommend$_{PL}$ & \textbf{the/these}$_{NOM-PL-F}$ & \textbf{authors} \\
  & \textit{The/These} & \textit{authors} & \textit{recommend} & \textit{the/this/a} & \textit{minister} \\
\hline
\end{tabular}}
\caption{Two examples of WOGLI premise-hypothesis pairs, one for the pattern \texttt{sing\_masc\_v\_sing\_masc} and one for the pattern  
\texttt{sing\_masc\_v\_pl\_fem}.
Underlined words have different surface forms in NE and E hypotheses and carry distinguishing morphological markers of case and/or number. Nouns belonging to the weak declension type are identified by \(^{\dagger}\).
Hypotheses H3 
are not part of WOGLI proper but will be used in a generalization set called WOGLI-OS-hard as they demand to both process marked OS word order as well as recognising non-entailment in the face of high word overlap.
As in the remainder of this paper, hypotheses with a marked word order are identified by an asterisk.}
\label{tab:wogli-examples}
\end{table*}

We have 17 patterns due to combinations of different argument NPs, including masculine and feminine proper names and common nouns as well as singular and plural arguments. 
Subjects/objects are either a simple proper name (such as \textit{Maria}) or  consist of an article\footnote{We used the articles \textit{ein} (indef.), \textit{der} (def.) and \textit{dieser} (demonstrative), as well as their feminine and plural forms.} and a common noun, e.g. \textit{der Arzt}/\textit{the doctor}. Consequently, each sentence always has a length of four or five words. 
A list of all  17 patterns is provided in Table
\ref{tab:patterns-list}
in the Appendix; we exclude the patterns in Table
\ref{tab:impossible-patterns}
in the Appendix as they generate ambiguous hypotheses, due to the absence of disambiguating morphological markers. The 17 patterns in WOGLI can be divided into two groups: 5 \textbf{all-singular} patterns that combine two singular nominal arguments (see first example in Table~\ref{tab:wogli-examples}) and 12 \textbf{singular-plural} patterns in which one argument is singular and the other one is plural (see second example in Table~\ref{tab:wogli-examples}). In all 9 patterns involving a masculine singular NP, (i) masculine determiners and (ii) masculine common nouns belonging to the weak declension type\footnote{The six masculine common nouns in WOGLI that belong to the weak declension type are \textit{Kunde}/\textit{Kunden}/\textit{client}, \textit{Student}/\textit{Studenten}/\textit{student}, \textit{Journalist}/\textit{Journalisten}/\textit{journalist}, \textit{Patient}/\textit{Patienten}/\textit{patient}, \textit{Soldat}/\textit{Soldaten}/\textit{soldier} and \textit{Zeuge}/\textit{Zeugen}/\textit{witness}. The remaining masculine nouns, e.g. \textit{Anwalt}/\textit{lawyer}, maintain the same surface forms in nominative and accusative.} carry morphological markers of case. Proper nouns never change surface forms. Additionally, in all \textbf{singular-plural} patterns, verb number agreement with the subject always leads to a change in the verb's surface form between E and NE hypotheses.

\paragraph{WOGLI statistics.} We generate 1,000 premises per pattern by randomly selecting  an appropriate subject/object and  verb from our lists, leading to 17,000 possible premises. As in random generation, some premises are generated twice, we deduplicate and are left with 16,971 premises.
H1-SO (NE) and H2-OS (E) are deterministically 
generated from the premises, leading to 33,942 sentence pairs.

All word lists with GXNLI-train frequencies and
translations can be found in our Github repository. Each of the 50 verb types appears between 308 and 383 times (mean: 339.4 times) in the 16,971  premises. They also appear 20 times on average per pattern in the premises. 
Table
\ref{tab:wogli-subj-obj-stats}
in the Appendix gives noun statistics for WOGLI.

\section{Experiments on WOGLI}\label{sec:results-wogli}

\paragraph{Models.} 



We use two German models  and one multilingual BERT model:
\begin{itemize}
    \item BERT-base\footnote{\url{https://huggingface.co/dbmdz/bert-base-german-cased}} is a cased base BERT model pre-trained by the MDZ Digital Library team on 16GB of German-language text.
    \item GBERT-large\footnote{\url{https://huggingface.co/deepset/gbert-large}} is a  BERT model
    pre-trained on 
        163.4GB of data \citep{chan-etal-2020-germans}, using the same cased vocabulary as BERT-base.\footnote{Other large-scale German language models such as Gott\-BERT \citep{scheible2020gottbert} and GELECTRA \citep{chan-etal-2020-germans} are of similar size and downstream performance. Thus, we use GBERT-large as a representative. }
    \item mBERT-base\footnote{\url{https://huggingface.co/bert-base-multilingual-cased}} is a cased BERT model  pre-trained on 104 languages \citep{devlin-etal-2019-bert}.
\end{itemize}
 
\noindent Since models were fine-tuned on GXNLI-train in a three-class setting, we merge contradiction and neutral into non-entailed predictions for evaluations on WOGLI. Fine-tuning 
details
are provided in Section
\ref{sec:appendix-fine-tuning}
of the Appendix.

\paragraph{Results (see Table~\ref{tab:results-on-xnli-and-wogli}).}
As a sanity check, we first test our models on GXNLI-test. 
Our models' performances on GXNLI-test are broadly in line with published work.
\citet{conneau-etal-2020-unsupervised} achieve an accuracy of 81.2\% on GXNLI-test with a monolingual BERT-base model, higher than our 76.67\%. However, their model uses a larger vocabulary (40k, ours: 31k) and was pre-trained on a larger corpus (up to 60GB, ours: 16GB). This particular model is unfortunately not available.
Other prior work concentrates on multilingual models:
GBERT-large's size is smaller than mT5-base (580m parameters)  by \citet{xue-etal-2021-mt5}, but its performance of 84.65\%  on GXNLI-test exceeds the one reported for  mT5-base (81.6\%). 
\citet{devlin-etal-2019-bert} achieve an accuracy of 75.9\% with mBERT-base (\textit{Translate Train Cased})\footnote{Results on XNLI are provided in the corresponding GitHub repository: \url{https://github.com/google-research/bert/blob/master/multilingual.md\#results}. 
}, in line with ours. 

On WOGLI, both base models
completely fail, labeling almost all instances as entailments. GBERT-large performs a bit better, suggesting that the language model's scale plays a role in its ability on WOGLI.
However, it still shows a strong tendency for the entailment class and the results are not robust across runs. 
Our vocabulary is frequent and present in GXNLI-train and 
our sentences have a very simple grammar.  Therefore, the models' poor performances on WOGLI 
suggest that not all German-specific linguistic phenomena are represented in the translated GXNLI-train, similar to our 
GXNLI word order analysis in Section~\ref{sec:background}.\footnote{One question that arises is whether even larger models or models pretrained on substantially more data will solve the problem. Other monolingual models for German are sparse.
We therefore ran two large, publically available,  multilingual model checkpoints  fine-tuned on  XNLI on WOGLI. They also do not perform well (see Section
\ref{sec:othermodels}
in the Appendix).}


\begin{table*}[!htb]
    \centering
    \begin{tabular}{llll}
        \hline
        Evaluation set & BERT-base (110m) & GBERT-large (335m) & mBERT-base (172m) \\
        \hline
        GXNLI-test & 76.65 (0.41) & 84.65 (0.163) & 75.16 (0.552) \\
        \hline
        WOGLI & 50.16 (0.133) & 57.68 (1.86) & 50.01 (0.015) \\
        WOGLI-SO (NE) & 0.33 (0.269) & 27.42 (7.828) & 0.02 (0.029)\\
        WOGLI-OS (E)* & 100 (0.005) & 87.94 (4.171) & 100 (0.0) \\
        \hline
    \end{tabular}
    \caption[Accuracies for two German and one multilingual model on GXNLI-test and WOGLI]{Accuracies for two German and one multilingual model on GXNLI-test and WOGLI, averaged over 5 runs. All are trained on GXNLI-train. 
    Accuracies are computed for 3 classes in GXNLI-test and 2 classes in WOGLI.}
    \label{tab:results-on-xnli-and-wogli}
\end{table*}


\section{Error analysis}\label{sec:error-analysis}

All analyses in this section are carried out on  ensemble predictions (majority vote of the 5 runs) of the strongest model in Table \ref{tab:results-on-xnli-and-wogli}, GBERT-large. 
The ensemble model reaches an accuracy of 57.82\% on WOGLI and 27.41\% on WOGLI-SO. 

\subsection{Fluency} \label{sec:error-analysis-fluency}

We measure the correlation of model performance and linguistic acceptability, approximating the latter via pseudo-loglikelihood \citep{salazar-etal-2020-masked}.
WOGLI premises have an average PLL of \(-30.54\) (SD: \(8.318\)). H1-SO (NE) hypotheses have an average PLL of \(-30.56\) (SD: \(8.287\)), while H2-OS (E) hypotheses are less fluent due to marked word order, with an average PLL of \(-36.53\) (SD: \(8.535\)). 
GBERT-large performs worse on SO (NE) pairs than on the less fluent OS (E) pairs; fluency thus does not play an important role in the model's performance on WOGLI. Instead, 
the lexical overlap heuristic \citep{naik-etal-2018-stress,mccoy-etal-2019-right,gururangan-etal-2018-annotation} is a possible reason for the degradation on non-entailed pairs.

\subsection{Performance by subject and object properties} \label{sec:error-analysis-properties}

We now focus on WOGLI-SO (NE)  only as this is the part of the dataset where the models fail.

\paragraph{Gender.} Regarding the gender of arguments in WOGLI, we formulate the following hypothesis:
\begin{itemize}
    \item[{\bf A1}] SO hypotheses with masculine subjects (objects) are easier to classify than the ones with feminine subjects (objects).
\end{itemize}
{\bf A1} can be explained by \textit{(a)} the presence of gender bias due to translation in GXNLI-train (see  Section \ref{sec:construction-wogli})  \textit{or} \textit{(b)} morphological differences between masculine and feminine NPs.

Performance on instances in  WOGLI-SO (NE) with masculine common noun subjects is indeed significantly higher than for 
feminine common noun subjects. The same holds for common noun objects (see also Table
\ref{tab:gender-probabilities}
in the Appendix). 
However, this does not transfer to proper names. 
 Gender bias in GXNLI-train \textit{(a)} as an explanation for {\bf A1} is therefore unlikely.

Morphological differences between feminine and masculine NPs  \textit{(b)}, however, are a possible explanation for {\bf A1}.
Feminine articles and common nouns have the same surface forms in accusative/nominative. Masculine articles and common nouns, however,  can bear morphological case markers. The masculine singular articles \textit{der}, \textit{ein} and \textit{dieser} are the only articles in WOGLI to change surface forms in the accusative to \textit{den}, \textit{einen} and \textit{diesen}. 
 Additionally,  singular masculine common nouns belonging to the weak declension type also carry case markers.
Morphological markers in some masculine NPs could thus be helpful for the model to distinguish subject from object.

\paragraph{Referential properties of subjects/objects.}

In prefield SO sentences, definite NPs tend to precede indefinite NPs \citep{weber2004word}, probably because indefinite 
constituents are often new and definite constituents are often given \citep{chafe1976}. Although XNLI and WOGLI 
do not contain  discourse context, preference for SO sentences with definite before indefinite NPs might
be encapsulated in pretraining data.
We thus hypothesize that:
\begin{itemize}
    \item[{\bf A2}] SO hypotheses in which a definite NP precedes an indefinite NP are easier to classify.
\end{itemize}

\noindent We separate WOGLI constituents into definite and indefinite following \citet{prince1992}: definite and demonstrative articles, as well as proper names are markers of definiteness, while indefinite articles point to indefiniteness. We then separate WOGLI-SO (16,971 pairs) into two groups: \textbf{preferred} (14,671 pairs) and \textbf{dispreferred} (2,300 pairs). Pairs in the \textbf{dispreferred} group are opposed to the aforementioned discourse hierarchy in that indefinite constituents precede definite constituents in the SO hypothesis. Pairs in the \textbf{preferred} group form the three other possible cases: definite precedes indefinite, definite precedes definite and indefinite precedes indefinite; these  cases are not in opposition to the hierarchy. 

GBERT-large achieves an accuracy of 29.85\% on the SO pairs in the \textbf{preferred} group but only
11.78\% on the \textbf{dispreferred} group (difference significant at 1\% significance level, z-test for proportions). Therefore we can confirm \textbf{A2}.



\paragraph{Number.} Lastly, we analyse the role of verb number agreement in classifying WOGLI-SO (NE) instances. As explained in Section~\ref{sec:construction-wogli}, WOGLI patterns either combine only singular arguments (\textbf{all-singular}) or a singular and a plural argument (\textbf{singular-plural}). Only in the latter group of patterns, subject-verb  agreement leads to a change in the verb's surface form from the premise to the H1-SO (NE) hypothesis (see \textit{empfehlen/recommend\(_{PL}\)} vs. \textit{empfiehlt/recommends\(_{SG}\)} in the second example in Table~\ref{tab:wogli-examples}).
We investigate the importance of verb number agreement for classifier performance by separating WOGLI-SO (16,971 pairs) into two groups, \textbf{all-singular} (4,997 pairs) and \textbf{singular-plural} (11,974 pairs).

GBERT-large achieves an accuracy of 36.66\% on the SO pairs in the \textbf{all-singular} group and 23.54\% on the \textbf{singular-plural} group (difference significant at 1\% significance level, z-test for proportions).
Thus the number switch in the verb occurring in \textbf{singular-plural} SO hypotheses is not a particularly helpful cue for the classifier.



\section{Data augmentation}\label{sec:data-aug-wogli}
Following  \citet{mccoy-etal-2019-right} and \citet{min-etal-2020-syntactic} on data augmentation with challenge sets, we hypothesize that augmenting GXNLI-train with a WOGLI subset can be helpful. 

We sample 1,037 premises and their corresponding E/NE hypotheses from WOGLI, resulting in 2,074 training instances. 
Each of the 17 patterns occurs 61 times.
All 50 verb lemmas are represented, each appearing between 18 and 25 times. All 181   noun forms appear at least once.\footnote{The nouns occur in varying frequencies due to the small size of the augmentation set.}


We concatenate these WOGLI  instances with GXNLI-train, name the resulting augmented training set GXNLI+1037 and shuffle it before fine-tuning GBERT-large 10 times on this augmented training set. We evaluate on the remaining 31,868 WOGLI instances, named WOGLI-test-1037. This augmented training set allows GBERT-large to classify WOGLI almost perfectly, while maintaining its performance on GXNLI-test (Table~\ref{tab:data-augmentation-results}).

\paragraph{Smaller augmentation size.} 
We fine-tune GBERT-large on a shuffled concatenation of GXNLI-train and only 102 WOGLI premises
sampled in a stratified manner from the aforementioned 1,037 premises 
along with both their corresponding NE and E hypotheses. Each one of the 17 patterns appears 6 times and each one of the 50 verb lemmas appears at least once and at most 4 times. Due to the small augmentation size, it is not possible to ensure representation of all 181 nouns, with 73 not appearing.
We evaluate on the remaining 33,738 WOGLI pairs, named WOGLI-test-102.
The smaller augmentation size yields a model that performs worse and less robustly on WOGLI test instances (Table~\ref{tab:data-augmentation-results}).

\begin{table*}[!htb]
    \centering
    \resizebox{\textwidth}{!}{
    \subfloat[Larger augmentation]{
    \begin{tabular}{lllll}
        \hline
        Evaluation set & GXNLI+1037 \\
        \hline
        GXNLI-test & 84.7 (0.301) \\
        WOGLI-test-1037 & 99.98 (0.016) \\
        WOGLI-SO-test-1037 (NE) & 99.99 (0.008) \\
        WOGLI-OS-test-1037 (E)* & 99.97 (0.03) \\
        \hline
    \end{tabular}}
    \quad
    \subfloat[Smaller augmentation]{
    \begin{tabular}{lllll}
        \hline
        Evaluation set & GXNLI+102 \\
        \hline
        GXNLI-test & 84.78 (0.244) \\
        WOGLI-test-102 & 86.04 (4.091) \\
        WOGLI-SO-test-102 (NE) & 87.57 (6.428) \\
        WOGLI-OS-test-102 (E)* & 84.52 (4.45) \\
        \hline
    \end{tabular}}}
    \caption[Accuracy for GBERT-large fine-tuned on GXNLI-train augmented with WOGLI instances]{Accuracy for GBERT-large fine-tuned on GXNLI-train augmented with WOGLI instances. Results are averaged over 10 runs and computed in a 3-class (GXNLI-test) or a 2-class (WOGLI-test) setting.} 
    \label{tab:data-augmentation-results}
\end{table*}

\section{Generalization experiments}\label{sec:generalization-wogli}

\citet{mccoy-etal-2019-right} investigate whether augmented 
models 
improved  by simply memorizing the seen templates. To do so, they  evaluate them on
pairs from unseen patterns.
Inspired by this setup, we study the models' generalization capabilities by evaluating them on four new evaluation sets that share structural and lexical similarities with the WOGLI pairs that were seen during fine-tuning. 

\subsection{Construction of generalization sets}

\paragraph{Pronoun subjects: WOGLI-p-subject.}
We replace the premise subject in WOGLI by a personal pronoun (\textit{He warns the client}).
Correspondingly, the  H1-SO (NE) hypothesis then has the pronoun as the object (\textit{The client warns him})  whereas the entailed H2-OS (E) hypothesis just  swaps word order with regards to the premise (\textit{The client warns he}) (see also Table
\ref{tab:wogli-generalization-examples-pronoun}
in the Appendix). 
To focus on the pronominalization change, the same 17 patterns, verb lemmas, proper nouns and common nouns are also used in WOGLI-p-subject. In addition to the previously mentioned morphological markers of case and/or verb number occurring in WOGLI sentences (Section~\ref{sec:construction-wogli}), the masculine singular pronoun \textit{er/he} (nominative) in WOGLI-p-subject changes surface form in the accusative case (\textit{ihn/him}). Feminine and plural pronouns (\textit{sie/she/her/they/them}) in WOGLI-p-subject, however, do not change surface form.
Some WOGLI premises can become duplicates after replacing the subject by a personal pronoun. Consider the two premises \textit{Die Ärzte warnen den Kunden}/\textit{The doctors$_{masc}$ warn the client} and \textit{Die Ärztinnen warnen den Kunden}/\textit{The doctors$_{fem}$ warn the client}. 
After replacing the subject, both premises lead to the new premise \textit{Sie warnen den Gast}/\textit{They warn the guest}, since plural masculine and plural feminine nominative personal pronouns have the same surface form in German. We  keep only one version for such duplicates. The new generalization set contains 13,802 unique premises, or a total of 27,604 pairs. 

\paragraph{Dative: WOGLI-dative.}
We collect a new  list of 22 transitive verbs that require dative  instead of accusative objects. All verbs are not symmetric, which ensures that NE hypotheses always have the correct gold label.
Each verb lemma appears at least 17 times in GXNLI-train. We use the same 
144 noun types as in WOGLI to generate  new instances. The premises again have  SVO structure, and H1 has SO (NE) and H2 has OS (E) structure. Therefore the instances are completely parallel to WOGLI apart
from the case of the object.

In these dative constructions, 24 patterns are possible (Table
\ref{tab:patterns-list}
in the Appendix). Each pattern appears 150 times in WOGLI-dative and each verb lemma appears between 132 and 182 times in the premises. All possible noun surface forms appear between 6 and 81 times in the premises. We generate 3,600 premises, or 7,200 pairs in total. Table 
\ref{tab:wogli-generalization-examples-dative} 
in the Appendix shows an example.
In WOGLI-dative, all determiners (singular and plural, feminine and masculine) change surface forms by case.
Additionally, plural masculine common nouns change surface forms in the dative if they do not end with -n in the nominative\footnote{Masculine nouns ending with -n in plural nominative are: \textit{Kunden/clients}, \textit{Professoren/professors}, \textit{Studenten/students}, \textit{Mentoren/mentors}, \textit{Patienten/patients}, \textit{Soldaten/soldiers}, \textit{Journalisten/journalists}, \textit{Zeugen/witnesses}. These nouns maintain the same surface forms in plural dative.}. As in WOGLI, singular masculine nouns of the weak declension type and verbs in singular-plural patterns also change surface forms.

\paragraph{Ditransitive verbs.}
We collect 21 ditransitive verbs (such as \textit{schicken/send} and \textit{verheimlichen/conceal}), each of which appears
at least 6 times in GXNLI-train. Verbs are grouped into 5 semantic categories (\texttt{giving, taking, sending, communication, secret}). 
Subjects and indirect objects of the verbs are compatible with the semantic class human, so that we can reuse the  144 noun types from WOGLI. For direct 
objects, we use a new list of 54 common nouns, appearing at least 15 times in GXNLI-train. They are grouped with the verb semantic categories so that resulting premises/hypotheses are meaningful (thus, you can combine the direct object \textit{Identit\"at/identity} with \texttt{secret} verbs but not with
\texttt{sending} verbs). The direct object is always preceded by a definite article.


Ditransitive premises follow the preferred word order SiO: subject-verb-IndirectObject-DirectObject (\textit{The waitresses give the merchant the cake}). Very similar to WOGLI, the not-entailed H1 hypothesis
swaps the underlying arguments of subject and indirect object, adapting case and number (\textit{The merchant give\underline{s} the waitresses the cake})
whereas the entailed H2 hypothesis reorders subject and indirect object into the marked iOS word order without changing the meaning by keeping case and number markers intact (\textit{The merchant giv\underline{e} the waitresses the cake}). 
The direct object is not affected.  An example is shown in Table
\ref{tab:wogli-generalization-examples-ditransitive}
in the Appendix.
With respect to morphological markers, WOGLI-ditransitive follows the same surface form changes between E and NE hypotheses as WOGLI-dative, since indirect objects in WOGLI-ditransitive are in dative case. 


Ditransitives 
allow for 24 unique patterns. 
We allow each pattern to appear 1,000 times leading to 12,000 premises or 24,000 pairs.


\paragraph{WOGLI-OS-hard (NE).}

Neither WOGLI  nor the previous generalization datasets contain instances where the marked OS word order leads to non-entailment, i.e. where you have to recognise non-entailment in the face of high word overlap while at the same time
processing a rare word order. 
Therefore we create a third hypothesis H3-OS (NE) for each WOGLI premise where we similar to 
H1 invert the underlying arguments but present this changed meaning in OS word order. Two examples
are given as H3-OS (NE) 
in Table~\ref{tab:wogli-examples}.  This is possible for all 17 WOGLI
patterns. Pairing H3 with each WOGLI premise leads to 16,971 new non-entailed  pairs.
All premises as well as all lexical items have been seen in normal WOGLI.


\subsection{Generalization results}

We evaluate GBERT-large fine-tuned on GXNLI without augmentation (GXNLI+0) as well as fine-tuned 
on GXNLI+1037 and on GXNLI+102
on our four generalization sets. 
Results are in Table~\ref{tab:generalization-results}.


GXNLI+1037 transfers very well to WOGLI-p-subject,
while GXNLI+102 reaches an accuracy of only 59.03\% on SO instances and is less robust. 
Thus, even for simple pronoun replacement a relatively large augmentation size is needed.
A similar picture emerges for WOGLI-dative. 
Since WOGLI-dative contains more patterns than WOGLI, we investigate whether GXNLI+102's poor performance is only observable in patterns that were not seen during fine-tuning but find no preference for seen or unseen patterns. 

With respect to ditransitive pairs, GXNLI+1037 has almost perfect accuracy and GXNLI+102 reaches its best generalization set performance, reaching similar results as on standard WOGLI.  

We hypothesized that generalization to H3-OS (NE) in WOGLI-OS-hard is the most difficult as it contains both marked word order and non-entailment, whereas  (i) in GXNLI, the marked word order
is 
very rare   (see Section~\ref{sec:background}) and (ii) 
in WOGLI, the marked word order
has always been seen with the entailment class, potentially tripping up an augmented model that could have learnt this hypothesis-only fact.
This turns out to be true:  GXNLI+0 classifies basically all WOGLI-OS-hard (NE) examples wrongly as entailment and performs even worse than the same model on the original WOGLI-SO (NE) non-entailed examples (see the 27.42\% in Table~\ref{tab:results-on-xnli-and-wogli}). 
With substantial augmentation (GXNLI+1037), performance is   slightly better but
the results are still both very low and   unstable. 

Our generalization experiments show that (i) the  augmentation set needs to be sufficiently large for  successful generalization to new NLI pairs that are structurally similar to WOGLI and (ii) models exposed to WOGLI  do not necessarily generalize well to some related datasets at all. As German word order is quite intricate  and  will have additional variations for embedded or non-declarative clauses this means training datasets need to be very large and varied to learn German word order. 

\begin{table*}[!htb]
    \centering
    \begin{tabular}{lllll}
        \hline
        Evaluation set & GXNLI+0 & GXNLI+102 & GXNLI+1037 \\
        \hline
        WOGLI-p-subject-test & 53.23 (1.715) & 77.97 (6.653) & 98.89 (0.957) \\
        WOGLI-p-subject-SO-test (NE) & 7.34 (4.16) & 59.03 (13.353) & 97.78 (1.916) \\
        WOGLI-p-subject-OS-test (E)* & 99.12 (0.74) & 96.91 (1.822) & 99.99 (0.008) \\
        \hline
        WOGLI-dative-test & 58.11 (2.789) & 79.4 (5.446) & 94.72 (0.565) \\
        WOGLI-dative-SO-test (NE) & 17.76 (6.223) & 60.87 (11.227) & 91.49 (1.421) \\
        WOGLI-dative-OS-test (E)* & 98.47 (0.776) & 97.93 (0.534) & 97.96 (0.504) \\
        \hline
        WOGLI-ditransitive-test & 73.93 (6.327) & 92.59 (4.634) & 99.58 (0.143) \\
        WOGLI-ditransitive-SiO-test (NE) & 50.23 (13.11) & 86.55 (9.41) & 99.62 (0.261) \\
        WOGLI-ditransitive-iOS-test (E)* & 97.63 (0.635) & 98.63 (0.591) & 99.55 (0.276) \\
        \hline
        WOGLI-OS-hard (NE)* & 0.15 (0.082) & 0.77 (0.75) & 23.45 (15.985)  \\
        \hline
    \end{tabular}
    \caption{Accuracy on generalization sets, averaged over 5 runs for GXNLI+0 and over 10 runs for remaining models} 
    \label{tab:generalization-results}
\end{table*}

\section{Related work}


Many English adversarial NLI datasets have been proposed. Some of these \citep{dasgupta2018,kim-etal-2018-teaching,nie2019analyzing,mccoy-etal-2019-right}, like us, include minimal pairs with a high word overlap between premise and hypotheses.  \citet{kim-etal-2018-teaching}, for example, change argument order to
generate non-entailments so that ``understanding'' word order is necessary to solve these.
However, in WOGLI, changes in argument order generate entailed \textit{and} non-entailed hypotheses,
depending on keeping or changing corresponding morphology. The more fixed English word order
does not allow for flexibility to that degree.
Regarding adversarial NLI datasets for German,
\citet{hartmann-etal-2021-multilingual} investigate  negation but do not
work on word order. 
\citet{tikhonova2022ad} propose NLI diagnostic datasets for French, German and Swedish. Sentence pairs are manually translated from the Russian TERRa dataset \citep{shavrina-etal-2020-russiansuperglue} as well as from the diagnostic dataset of GLUE \citep{wang-etal-2018-glue}. 
We inspected a random 100 hypotheses of the German TERRa dataset,
none of which were in marked word order. The translated GLUE benchmark
is annotated with linguistic features relevant for entailment such as  lexical semantics, logic,  and predicate-argument structure. 
Only the predicate-argument structure examples include  a handful  where word order of arguments has been inverted between premise and hypothesis. However, resulting hypotheses were often ambiguous
and --- in our opinion --- wrongly annotated as not-entailed.
Consider the premise \textit{John zerbrach das Fenster}/\textit{John broke the window} and the hypothesis \textit{Das Fenster hat John eingeschlagen}, which is ambiguous between \textit{The window$_{NOM}$ broke John$_{ACC}$} (SO order, NE) OR \textit{The window$_{ACC}$ broke John$_{NOM}$} (OS order, E). 
This is annotated as non-entailment  in the dataset, assuming SO order with an implausible semantic reading, whereas the marked word order with a plausible semantic reading leads to entailment.


Unlike us, both  datasets do not emphasise word order.
They are also based on translations   and therefore rarely contain OS hypotheses.

\section{Conclusion}
We created WOGLI, a new NLI challenge set, in order to examine the challenges brought by the freer German word order. Premises, entailed and not-entailed hypotheses contain exactly the same lemmata; the two hypotheses differ only in word order and morphological changes but change label.
Three current BERT-based models fine-tuned on GXNLI-train struggle on WOGLI pairs. This poor performance mirrors the fact that translated NLI training sets such as GXNLI do not incorporate all required linguistic phenomena that are specific to the target language, German. 
We find that the number of WOGLI pairs for augmentation during fine-tuning must be sufficiently high in order to (i) learn WOGLI and (ii) generalize to other WOGLI-like pairs. Even with a larger augmentation set and  a large pretrained model, a generalization set that differs more from WOGLI, such as WOGLI-OS-hard (NE) , remains difficult. 

In future experiments,
we will expand WOGLI datasets to contain additional variation, such as tense variation, more complex sentence structure (additional arguments and adjuncts, active/passive), more complex constituent structure 
and  other sentence types (non-declarative, embedded).
This   will also allow us to conduct more fine-grained error analyses regarding the hierarchies that influence the linearization of arguments and thus word order.


\section*{Acknowledgments}
Ines Reinig is funded by the German Research Foundation (DFG) under the UNCOVER project (PO 1900/7-1 and RE 3536/3-1). 
\newpage
\bibliographystyle{acl_natbib}
\bibliography{anthology,acl2021}

\begin{thebibliography}{31}
\expandafter\ifx\csname natexlab\endcsname\relax\def\natexlab#1{#1}\fi

\bibitem[{Bader et~al.(2017)Bader, Ellsiepen, Koukoulioti, and
  Portele}]{bader2017filling}
Markus Bader, Emilia Ellsiepen, Vasiliki Koukoulioti, and Yvonne Portele. 2017.
\newblock Filling the prefield: Findings and challenges.
\newblock In \emph{DGfS 2016 workshop “V2 in grammar and processing: Its
  causes and its consequences”}, pages 27--49.

\bibitem[{Bader and H{\"a}ussler(2010)}]{bader2010word}
Markus Bader and Jana H{\"a}ussler. 2010.
\newblock Word order in {G}erman: {A} corpus study.
\newblock \emph{Lingua}, 120(3):717--762.

\bibitem[{Bader and Portele(2019)}]{bader2019givenness}
Markus Bader and Yvonne Portele. 2019.
\newblock Givenness and the licensing of object-first order in {G}erman: The
  effect of referential form.
\newblock In \emph{Proceedings of Linguistic Evidence 2018: Experimental Data
  Drives Linguistic Theory}. Universit{\"a}t T{\"u}bingen.

\bibitem[{Brown et~al.(2020)Brown, Mann, Ryder, Subbiah, Kaplan, Dhariwal,
  Neelakantan, Shyam, Sastry, Askell et~al.}]{brown2020language}
Tom Brown, Benjamin Mann, Nick Ryder, Melanie Subbiah, Jared~D Kaplan, Prafulla
  Dhariwal, Arvind Neelakantan, Pranav Shyam, Girish Sastry, Amanda Askell,
  et~al. 2020.
\newblock Language models are few-shot learners.
\newblock \emph{Advances in neural information processing systems},
  33:1877--1901.

\bibitem[{Chafe(1976)}]{chafe1976}
Wallace Chafe. 1976.
\newblock \emph{Givenness, constrastiveness, definiteness, subjects, topics,
  and point of view}, pages 25--55. Academic Press, New York.

\bibitem[{Chan et~al.(2020)Chan, Schweter, and
  M{\"o}ller}]{chan-etal-2020-germans}
Branden Chan, Stefan Schweter, and Timo M{\"o}ller. 2020.
\newblock \href {https://www.aclweb.org/anthology/2020.coling-main.598}
  {{G}erman{'}s next language model}.
\newblock In \emph{Proceedings of the 28th International Conference on
  Computational Linguistics}, pages 6788--6796, Barcelona, Spain (Online).
  International Committee on Computational Linguistics.

\bibitem[{Conneau et~al.(2020)Conneau, Khandelwal, Goyal, Chaudhary, Wenzek,
  Guzm{\'a}n, Grave, Ott, Zettlemoyer, and
  Stoyanov}]{conneau-etal-2020-unsupervised}
Alexis Conneau, Kartikay Khandelwal, Naman Goyal, Vishrav Chaudhary, Guillaume
  Wenzek, Francisco Guzm{\'a}n, Edouard Grave, Myle Ott, Luke Zettlemoyer, and
  Veselin Stoyanov. 2020.
\newblock \href {https://doi.org/10.18653/v1/2020.acl-main.747} {Unsupervised
  cross-lingual representation learning at scale}.
\newblock In \emph{Proceedings of the 58th Annual Meeting of the Association
  for Computational Linguistics}, pages 8440--8451, Online. Association for
  Computational Linguistics.

\bibitem[{Conneau et~al.(2018)Conneau, Rinott, Lample, Williams, Bowman,
  Schwenk, and Stoyanov}]{conneau-etal-2018-xnli}
Alexis Conneau, Ruty Rinott, Guillaume Lample, Adina Williams, Samuel Bowman,
  Holger Schwenk, and Veselin Stoyanov. 2018.
\newblock \href {https://doi.org/10.18653/v1/D18-1269} {{XNLI}: Evaluating
  cross-lingual sentence representations}.
\newblock In \emph{Proceedings of the 2018 Conference on Empirical Methods in
  Natural Language Processing}, pages 2475--2485, Brussels, Belgium.
  Association for Computational Linguistics.

\bibitem[{Dasgupta et~al.(2018)Dasgupta, Guo, Stuhlm{\"u}ller, Gershman, and
  Goodman}]{dasgupta2018}
Ishita Dasgupta, Demi Guo, Andreas Stuhlm{\"u}ller, Samuel~J. Gershman, and
  Noah~D. Goodman. 2018.
\newblock Evaluating compositionality in sentence embeddings.
\newblock In \emph{Proceedings of the Fortieth Annual Conference of the
  Cognitive Science Society}.

\bibitem[{Devlin et~al.(2019)Devlin, Chang, Lee, and
  Toutanova}]{devlin-etal-2019-bert}
Jacob Devlin, Ming-Wei Chang, Kenton Lee, and Kristina Toutanova. 2019.
\newblock \href {https://doi.org/10.18653/v1/N19-1423} {{BERT}: Pre-training of
  deep bidirectional transformers for language understanding}.
\newblock In \emph{Proceedings of the 2019 Conference of the North {A}merican
  Chapter of the Association for Computational Linguistics: Human Language
  Technologies, Volume 1 (Long and Short Papers)}, pages 4171--4186,
  Minneapolis, Minnesota. Association for Computational Linguistics.

\bibitem[{Drach(1937)}]{drach_1937}
Erich Drach. 1937.
\newblock \emph{Grundgedanken der Deutschen Satzlehre}.
\newblock M. Diesterweg.

\bibitem[{D{\"u}rscheid(2012)}]{durscheid2012syntax}
Christa D{\"u}rscheid. 2012.
\newblock \emph{Syntax: Grundlagen und Theorien}, volume~6.
\newblock Vandenhoeck \& Ruprecht.

\bibitem[{Gururangan et~al.(2018)Gururangan, Swayamdipta, Levy, Schwartz,
  Bowman, and Smith}]{gururangan-etal-2018-annotation}
Suchin Gururangan, Swabha Swayamdipta, Omer Levy, Roy Schwartz, Samuel Bowman,
  and Noah~A. Smith. 2018.
\newblock \href {https://doi.org/10.18653/v1/N18-2017} {Annotation artifacts in
  natural language inference data}.
\newblock In \emph{Proceedings of the 2018 Conference of the North {A}merican
  Chapter of the Association for Computational Linguistics: Human Language
  Technologies, Volume 2 (Short Papers)}, pages 107--112, New Orleans,
  Louisiana. Association for Computational Linguistics.

\bibitem[{Hartmann et~al.(2021)Hartmann, de~Lhoneux, Hershcovich,
  Kementchedjhieva, Nielsen, Qiu, and
  S{\o}gaard}]{hartmann-etal-2021-multilingual}
Mareike Hartmann, Miryam de~Lhoneux, Daniel Hershcovich, Yova Kementchedjhieva,
  Lukas Nielsen, Chen Qiu, and Anders S{\o}gaard. 2021.
\newblock \href {https://doi.org/10.18653/v1/2021.conll-1.19} {A multilingual
  benchmark for probing negation-awareness with minimal pairs}.
\newblock In \emph{Proceedings of the 25th Conference on Computational Natural
  Language Learning}, pages 244--257, Online. Association for Computational
  Linguistics.

\bibitem[{Hu et~al.(2020)Hu, Ruder, Siddhant, Neubig, Firat, and
  Johnson}]{hu2020xtreme}
Junjie Hu, Sebastian Ruder, Aditya Siddhant, Graham Neubig, Orhan Firat, and
  Melvin Johnson. 2020.
\newblock Xtreme: A massively multilingual multi-task benchmark for evaluating
  cross-lingual generalisation.
\newblock In \emph{International Conference on Machine Learning}, pages
  4411--4421. PMLR.

\bibitem[{Kim et~al.(2018)Kim, Malon, and Kadav}]{kim-etal-2018-teaching}
Juho Kim, Christopher Malon, and Asim Kadav. 2018.
\newblock \href {https://doi.org/10.18653/v1/W18-5512} {Teaching syntax by
  adversarial distraction}.
\newblock In \emph{Proceedings of the First Workshop on Fact Extraction and
  {VER}ification ({FEVER})}, pages 79--84, Brussels, Belgium. Association for
  Computational Linguistics.

\bibitem[{McCoy et~al.(2019)McCoy, Pavlick, and Linzen}]{mccoy-etal-2019-right}
Tom McCoy, Ellie Pavlick, and Tal Linzen. 2019.
\newblock \href {https://doi.org/10.18653/v1/P19-1334} {Right for the wrong
  reasons: Diagnosing syntactic heuristics in natural language inference}.
\newblock In \emph{Proceedings of the 57th Annual Meeting of the Association
  for Computational Linguistics}, pages 3428--3448, Florence, Italy.
  Association for Computational Linguistics.

\bibitem[{Min et~al.(2020)Min, McCoy, Das, Pitler, and
  Linzen}]{min-etal-2020-syntactic}
Junghyun Min, R.~Thomas McCoy, Dipanjan Das, Emily Pitler, and Tal Linzen.
  2020.
\newblock \href {https://doi.org/10.18653/v1/2020.acl-main.212} {Syntactic data
  augmentation increases robustness to inference heuristics}.
\newblock In \emph{Proceedings of the 58th Annual Meeting of the Association
  for Computational Linguistics}, pages 2339--2352, Online. Association for
  Computational Linguistics.

\bibitem[{Naik et~al.(2018)Naik, Ravichander, Sadeh, Rose, and
  Neubig}]{naik-etal-2018-stress}
Aakanksha Naik, Abhilasha Ravichander, Norman Sadeh, Carolyn Rose, and Graham
  Neubig. 2018.
\newblock \href {https://www.aclweb.org/anthology/C18-1198} {Stress test
  evaluation for natural language inference}.
\newblock In \emph{Proceedings of the 27th International Conference on
  Computational Linguistics}, pages 2340--2353, Santa Fe, New Mexico, USA.
  Association for Computational Linguistics.

\bibitem[{Nie et~al.(2019)Nie, Wang, and Bansal}]{nie2019analyzing}
Yixin Nie, Yicheng Wang, and Mohit Bansal. 2019.
\newblock Analyzing compositionality-sensitivity of {NLI} models.
\newblock In \emph{Proceedings of the AAAI Conference on Artificial
  Intelligence}, volume~33, pages 6867--6874.

\bibitem[{Prince(1992)}]{prince1992}
Ellen~F Prince. 1992.
\newblock The {ZPG} letter: Subjects, definiteness, and information-status.
\newblock \emph{Discourse description: diverse analyses of a fund raising
  text}, pages 295--325.

\bibitem[{Salazar et~al.(2020)Salazar, Liang, Nguyen, and
  Kirchhoff}]{salazar-etal-2020-masked}
Julian Salazar, Davis Liang, Toan~Q. Nguyen, and Katrin Kirchhoff. 2020.
\newblock \href {https://doi.org/10.18653/v1/2020.acl-main.240} {Masked
  language model scoring}.
\newblock In \emph{Proceedings of the 58th Annual Meeting of the Association
  for Computational Linguistics}, pages 2699--2712, Online. Association for
  Computational Linguistics.

\bibitem[{Scheible et~al.(2020)Scheible, Thomczyk, Tippmann, Jaravine, and
  Boeker}]{scheible2020gottbert}
Raphael Scheible, Fabian Thomczyk, Patric Tippmann, Victor Jaravine, and Martin
  Boeker. 2020.
\newblock Gott{BERT}: a pure {G}erman language model.
\newblock \emph{arXiv preprint arXiv:2012.02110}.

\bibitem[{Shavrina et~al.(2020)Shavrina, Fenogenova, Anton, Shevelev, Artemova,
  Malykh, Mikhailov, Tikhonova, Chertok, and
  Evlampiev}]{shavrina-etal-2020-russiansuperglue}
Tatiana Shavrina, Alena Fenogenova, Emelyanov Anton, Denis Shevelev, Ekaterina
  Artemova, Valentin Malykh, Vladislav Mikhailov, Maria Tikhonova, Andrey
  Chertok, and Andrey Evlampiev. 2020.
\newblock \href {https://doi.org/10.18653/v1/2020.emnlp-main.381}
  {{R}ussian{S}uper{GLUE}: A {R}ussian language understanding evaluation
  benchmark}.
\newblock In \emph{Proceedings of the 2020 Conference on Empirical Methods in
  Natural Language Processing (EMNLP)}, pages 4717--4726, Online. Association
  for Computational Linguistics.

\bibitem[{Siewierska(1993)}]{siewierska1993}
Anna Siewierska. 1993.
\newblock \href {https://doi.org/doi:10.1515/9783110095869.1.13.826} {On the
  interplay of factors in the determination of word order}.
\newblock In Joachim Jacobs, Arnim von Stechow, Wolfgang Sternefeld, and Theo
  Vennemann, editors, \emph{An International Handbook of Contemporary
  Research}, pages 826--846. De Gruyter Mouton, Berlin • New York.

\bibitem[{Suzgun et~al.(2022)Suzgun, Melas-Kyriazi, and
  Jurafsky}]{suzgun2022prompt}
Mirac Suzgun, Luke Melas-Kyriazi, and Dan Jurafsky. 2022.
\newblock Prompt-and-rerank: A method for zero-shot and few-shot arbitrary
  textual style transfer with small language models.
\newblock \emph{arXiv preprint arXiv:2205.11503}.

\bibitem[{Tikhonova et~al.(2022)Tikhonova, Mikhailov, Pisarevskaya, Malykh, and
  Shavrina}]{tikhonova2022ad}
Maria Tikhonova, Vladislav Mikhailov, Dina Pisarevskaya, Valentin Malykh, and
  Tatiana Shavrina. 2022.
\newblock Ad astra or astray: Exploring linguistic knowledge of multilingual
  {BERT} through {NLI} task.
\newblock \emph{Natural Language Engineering}, pages 1--30.

\bibitem[{Wang et~al.(2018)Wang, Singh, Michael, Hill, Levy, and
  Bowman}]{wang-etal-2018-glue}
Alex Wang, Amanpreet Singh, Julian Michael, Felix Hill, Omer Levy, and Samuel
  Bowman. 2018.
\newblock \href {https://doi.org/10.18653/v1/W18-5446} {{GLUE}: A multi-task
  benchmark and analysis platform for natural language understanding}.
\newblock In \emph{Proceedings of the 2018 {EMNLP} Workshop {B}lackbox{NLP}:
  Analyzing and Interpreting Neural Networks for {NLP}}, pages 353--355,
  Brussels, Belgium. Association for Computational Linguistics.

\bibitem[{Weber and M{\"u}ller(2004)}]{weber2004word}
Andrea Weber and Karin M{\"u}ller. 2004.
\newblock Word order variation in {G}erman main clauses: {A} corpus analysis.
\newblock In \emph{20th International Conference on Computational Linguistics}.

\bibitem[{Williams et~al.(2018)Williams, Nangia, and
  Bowman}]{williams-etal-2018-broad}
Adina Williams, Nikita Nangia, and Samuel Bowman. 2018.
\newblock \href {https://doi.org/10.18653/v1/N18-1101} {A broad-coverage
  challenge corpus for sentence understanding through inference}.
\newblock In \emph{Proceedings of the 2018 Conference of the North {A}merican
  Chapter of the Association for Computational Linguistics: Human Language
  Technologies, Volume 1 (Long Papers)}, pages 1112--1122, New Orleans,
  Louisiana. Association for Computational Linguistics.

\bibitem[{Xue et~al.(2021)Xue, Constant, Roberts, Kale, Al-Rfou, Siddhant,
  Barua, and Raffel}]{xue-etal-2021-mt5}
Linting Xue, Noah Constant, Adam Roberts, Mihir Kale, Rami Al-Rfou, Aditya
  Siddhant, Aditya Barua, and Colin Raffel. 2021.
\newblock \href {https://doi.org/10.18653/v1/2021.naacl-main.41} {m{T}5: A
  massively multilingual pre-trained text-to-text transformer}.
\newblock In \emph{Proceedings of the 2021 Conference of the North American
  Chapter of the Association for Computational Linguistics: Human Language
  Technologies}, pages 483--498, Online. Association for Computational
  Linguistics.

\end{thebibliography}

\newpage
\appendix













\appendix

\section{SVO patterns}
\label{sec:appendix-patterns}

Table~\ref{tab:patterns-list} lists the patterns that we use to build WOGLI and generalization sets.
Table~\ref{tab:impossible-patterns} lists the 8 patterns that we exclude from WOGLI. The noun phrases in the SO and OS form have the same morphological surface form and the verb also has the same form in both word orders. Therefore, SO and OS meaning are not distinguishable.

\begin{table*}[!htb]
    \centering
    \begin{tabular}{ll}
\hline
Dative, Ditransitive & WOGLI, WOGLI-OS-hard (NE) \\
\hline
\texttt{pnoun\_v\_sing\_masc}  & \texttt{pnoun\_v\_sing\_masc} \\
\texttt{pnoun\_v\_plural\_masc}  & \texttt{pnoun\_v\_plural\_masc} \\
\texttt{pnoun\_v\_plural\_fem} & \texttt{pnoun\_v\_plural\_fem} \\
\texttt{pnoun\_v\_sing\_fem} & \\
\texttt{plural\_masc\_v\_pnoun} & \texttt{plural\_masc\_v\_pnoun} \\
\texttt{plural\_masc\_v\_sing\_masc} & \texttt{plural\_masc\_v\_sing\_masc} \\
\texttt{plural\_masc\_v\_sing\_fem} & \texttt{plural\_masc\_v\_sing\_fem} \\
\texttt{plural\_masc\_v\_plural\_fem} &  \\
\texttt{plural\_masc\_v\_plural\_masc} &  \\
\texttt{plural\_fem\_v\_sing\_masc} & \texttt{plural\_fem\_v\_sing\_masc}\\
\texttt{plural\_fem\_v\_sing\_fem} & \texttt{plural\_fem\_v\_sing\_fem}\\
\texttt{plural\_fem\_v\_pnoun} & \texttt{plural\_fem\_v\_pnoun} \\
\texttt{plural\_fem\_v\_plural\_fem} &   \\
\texttt{plural\_fem\_v\_plural\_masc} &   \\
\texttt{sing\_masc\_v\_sing\_masc} & \texttt{sing\_masc\_v\_sing\_masc} \\
\texttt{sing\_masc\_v\_plural\_masc} & \texttt{sing\_masc\_v\_plural\_masc} \\
\texttt{sing\_masc\_v\_plural\_fem} & \texttt{sing\_masc\_v\_plural\_fem} \\
\texttt{sing\_masc\_v\_sing\_fem} & \texttt{sing\_masc\_v\_sing\_fem} \\
\texttt{sing\_masc\_v\_pnoun} & \texttt{sing\_masc\_v\_pnoun} \\
\texttt{sing\_fem\_v\_sing\_masc} & \texttt{sing\_fem\_v\_sing\_masc} \\
\texttt{sing\_fem\_v\_plural\_fem} & \texttt{sing\_fem\_v\_plural\_fem} \\
\texttt{sing\_fem\_v\_plural\_masc} & \texttt{sing\_fem\_v\_plural\_masc} \\
\texttt{sing\_fem\_v\_pnoun} &  \\
\texttt{sing\_fem\_v\_sing\_fem} & \\
\hline
    \end{tabular}
    \caption{Exhaustive list of patterns used to build WOGLI-dative, WOGLI-ditransitive (24 patterns) and WOGLI (17 patterns). WOGLI-OS-hard (NE) uses the same patterns as WOGLI.}
    \label{tab:patterns-list}
\end{table*}

\begin{table*}[!htb]
    \centering
    \resizebox{\textwidth}{!}{
    \begin{tabular}{lp{6.5cm}p{6.5cm}l}
        \hline
        Pattern & Premise & Hypothesis & Label \\
        \hline
        \texttt{sing\_fem\_v\_pnoun} & Die Freundin begr\"u\ss{}t David . & David begr\"u\ss{}t die Freundin . & ? \\
         & The friend$_{CASE?-SING-FEM}$ greets David$_{CASE?-SING-MASC}$ & David$_{CASE?-SING-MASC}$ greets the friend$_{CASE?-SING-FEM}$ \\
        \hline
        \texttt{pnoun\_v\_sing\_fem} & David begr\"u\ss{}t die Freundin . & Die Freundin begr\"u\ss{}t David . & ? \\
         & David$_{CASE?-SING-MASC}$ greets the friend$_{CASE?-SING-FEM}$ & The friend$_{CASE?-SING-FEM}$ greets David$_{CASE?-SING-MASC}$ \\
        \hline
        \texttt{pnoun\_v\_pnoun} & Walter begr\"u\ss{}t David . & David begr\"u\ss{}t Walter . & ? \\
         & Walter$_{CASE?-SING-MASC}$ greets David$_{CASE?-SING-MASC}$ & David$_{CASE?-SING-MASC}$ greets Walter$_{CASE?-SING-MASC}$ \\
        \hline
        \texttt{sing\_fem\_v\_sing\_fem} & Die Mitbewohnerin begr\"u\ss{}t die Freundin . & Die Freundin begr\"u\ss{}t die Mitbewohnerin . & ? \\
         & The flatmate$_{CASE?-SING-FEM}$ greets the friend$_{CASE?-SING-FEM}$ & The friend$_{CASE?-SING-FEM}$ greets the flatmate$_{CASE?-SING-FEM}$ \\
        \hline
        \texttt{plural\_fem\_v\_plural\_fem} & Die Freundinnen begr\"u\ss{}en die Mitbewohnerinnen . & Die Mitbewohnerinnen begr\"u\ss{}en die Freundinnen . & ? \\
         & The friends$_{CASE?-PL-FEM}$ greet the flatmates$_{CASE?-PL-FEM}$ & The flatmates$_{CASE?-PL-FEM}$ greet the friends$_{CASE?-PL-FEM}$ \\
        \hline
        \texttt{plural\_masc\_v\_plural\_masc} & Die Freunde begr\"u\ss{}en die Mitbewohner . & Die Mitbewohner begr\"u\ss{}en die Freunde . & ? \\
         & The friends$_{CASE?-PL-MASC}$ greet the flatmates$_{CASE?-PL-MASC}$ & The flatmates$_{CASE?-PL-MASC}$ greet the friends$_{CASE?-PL-MASC}$ \\
        \hline
        \texttt{plural\_masc\_v\_plural\_fem} & Die Freunde begr\"u\ss{}en die Mitbewohnerinnen . & Die Mitbewohnerinnen begr\"u\ss{}en die Freunde . & ? \\
         & The friends$_{CASE?-PL-MASC}$ greet the flatmates$_{CASE?-PL-FEM}$ & The flatmates$_{CASE?-PL-FEM}$ greet the friends$_{CASE?-PL-MASC}$ \\
        \hline
        \texttt{plural\_fem\_v\_plural\_masc} & Die Freundinnen begr\"u\ss{}en die Mitbewohner . & Die Mitbewohner begr\"u\ss{}en die Freundinnen . & ? \\
         & The friends$_{CASE?-PL-FEM}$ greet the flatmates$_{CASE?-PL-MASC}$ & The flatmates$_{CASE?-PL-MASC}$ greet the friends$_{CASE?-PL-FEM}$ \\
        \hline
    \end{tabular}}
    \caption{The 8 patterns that we exclude from WOGLI and WOGLI-OS-hard (NE)}
    \label{tab:impossible-patterns}
\end{table*}

\section{WOGLI statistics}

Table \ref{tab:wogli-subj-obj-stats} shows counts and ratios for the subject and object roles in WOGLI. The different noun categories shown in the first column generally take the subject role as often as they take the object role.


\begin{table*}[!htb]
    \centering
    \resizebox{\textwidth}{!}{
    \begin{tabular}{p{1.1cm}r@{\hskip 0.5in}p{1cm}rr@{\hskip 0.5in}p{1cm}rr@{\hskip 0.5in}rrrr}
        \hline
        Noun & \# types & \multicolumn{3}{l}{Subject count} & \multicolumn{3}{l}{Object count} & \multicolumn{4}{l}{Subject/Object ratio} \\
         & & Mean & Min & Max & Mean & Min & Max & Mean & Median & Min & Max \\
        \hline
        Masc pnoun & 41 & 36.8 \small{(4.22)} & 27 & 49 & 36.5 \small{(6.89)} & 25 & 53 & 1.05 & 1.0 & 0.58 & 1.72 \\
        \hline
        Fem pnoun & 41 & 36.3 \small{(5.71)} & 27 & 52 &  36.5 \small{(6.18)} & 23 & 49 & 1.02 & 1.0 & 0.63 & 1.57 \\
        \hline
        Masc sing. cnoun & 38 & 131.4 \small{(12.23)} & 108 & 163 & 131.5 \small{(10.38)} & 114 & 158 & 1.01 & 1.0 & 0.78 & 1.38 \\
        \hline
        Masc pl. cnoun & 38 & 78.7 \small{(9.23)} & 60 & 97 & 78.8 \small{(8.86)} & 53 & 99 & 1.01 & 1.01 & 0.75 & 1.36 \\
        \hline
        Fem sing. cnoun & 24 & 124.8 \small{(11.05)} & 103 & 143 & 124.7 \small{(7.90)} & 109 & 137 & 1.01 & 1.0 & 0.81 & 1.25 \\
        \hline
        Fem pl. cnoun & 24 & 124.7 \small{(12.08)} & 100 & 144 & 124.8 \small{(12.40)} & 101 & 145 & 1.01 & 0.98 & 0.77 & 1.29 \\
        \hline
    \end{tabular}}
    \caption[Average counts and subject to object ratios for different groups of nouns in WOGLI]{Average counts and subject to object ratios for different groups of nouns in WOGLI. For example, masculine proper nouns are subjects 36.8 times and objects 36.5 times on average. Values in parentheses are standard deviations.}
    \label{tab:wogli-subj-obj-stats}
\end{table*}



\section{Fine-tuning details} \label{sec:appendix-fine-tuning}
The input sequence, consisting of the premise-hypothesis pair, is encoded using the given BERT model. The final hidden state of the special \texttt{[CLS]} token constitutes the aggregate representation of the input sequence, following \citet{devlin-etal-2019-bert}. This representation is then passed through a dropout layer and a linear classification layer, which maps it to the three-label classification space. 
All models were fine-tuned for three epochs, with linear warmup over 6\% of the first steps and a maximum sequence length of 128.
BERT-base and mBERT-base were fine-tuned with a batch size of 16 and a learning rate of \(5e^{-5}\). GBERT-large was fine-tuned with a batch size of 32 and a learning rate of \(5e^{-6}\).
Regarding mBERT-base, fine-tuning on the translated training set is comparable to the \textit{Translate Train} setup in \citet{conneau-etal-2018-xnli}. We also experimented with fine-tuning mBERT-base on the English MNLI training set, similarly to the \textit{Zero Shot} setup in \citet{conneau-etal-2018-xnli}, but found better validation set accuracy using the translated training set.
GBERT-large was fine-tuned on one NVIDIA A6000 GPU. Base models were fine-tuned on one NVIDIA T4 GPU. 
Fine-tuning took approximately 2 hours per run.

\section{WOGLI results for other models}\label{sec:othermodels}

Table \ref{tab:results-other-models} shows results on WOGLI for publically available checkpoints (single runs) of two larger multilingual models: XLM-RoBERTa-large\footnote{\url{https://huggingface.co/joeddav/xlm-roberta-large-xnli}} \citep{conneau-etal-2020-unsupervised} (XLM-R) and the generative encoder-decoder model mT5-large\footnote{\url{https://huggingface.co/alan-turing-institute/mt5-large-finetuned-mnli-xtreme-xnli}} \citep{xue-etal-2021-mt5}. These two models have considerably more parameters than GBERT-large.
According to the respective model cards: 
\begin{itemize}
    \item XLM-R was fine-tuned on the concatenation of the MNLI training set and the XNLI validation and test sets.
    \item mT5-large was fine-tuned on the MNLI and the XTREME XNLI\footnote{This version of the XNLI dataset contains different machine translations than the original XNLI dataset: \url{https://www.tensorflow.org/datasets/catalog/xtreme_xnli}} \citep{hu2020xtreme} training sets.
\end{itemize}

Both large models perform much better than the two base models (see 
Table~3 in the paper), which suggests again that model scale is relevant on WOGLI. However, they do not achieve higher overall accuracies than GBERT-large (average: 57.68\%).
Interestingly, mT5-large performs best on WOGLI-SO, but struggles substantially on WOGLI-OS, often labeling these pairs as non-entailments. 

\begin{table*}[!htb]
    \centering
    \begin{tabular}{lll}
        \hline
        Evaluation set & XLM-R (550m) & mT5-large (1.2b)\\
        \hline
        WOGLI & 55.42 & 52.26 \\
        WOGLI-SO (NE) & 46.2 & 68.7 \\
        WOGLI-OS (E)* & 64.64 & 35.82 \\ 
        \hline
    \end{tabular}
    \caption{Accuracies on WOGLI for two larger multilingual models. Results are for single runs.}
    \label{tab:results-other-models}
\end{table*}

\paragraph{ChatGPT: discussion.} In a small-scale experiment, we evaluate the ability of the recently made available research preview for the chatbot ChatGPT (February 13 version) by OpenAI\footnote{\url{https://openai.com/blog/chatgpt}} on WOGLI. This chatbot is based on the autoregressive GPT-3 model \citep{brown2020language}, as opposed to autoencoding models such as BERT, and has recently drawn a lot of attention in the AI community. We attempted to obtain classifications from ChatGPT on a WOGLI subset consisting of of 51 WOGLI-SO and 51 WOGLI-OS pairs. 
However, we observed (i) a strong prompt-dependence \citep{suzgun2022prompt}, as even minor changes in the prompt's phrasing lead to different answers by the chatbot and (ii) overall inconsistent results across multiple instances of showing the model the same sets of pairs. Due to the inconsistency of these preliminary results, we leave it to future work to assess ChatGPT's capabilities on WOGLI in a more systematic manner and for a range of different prompt styles.

\section{Error analysis: performance by gender}

Table \ref{tab:gender-probabilities} provides more detailed results for the analysis discussed in 
Section~5.2.

\begin{table*}[!htb]
    \centering
    \begin{tabular}{llllrr}
        \hline
        Constituent & \multicolumn{2}{l}{Argument} & \multicolumn{2}{l}{Conditional probability} & Signif. \\
        \cline{4-5}
         & Premise & Hypo (SO) & Definition & Value (\%) & \\
        \hline
        \multirow{ 2}{*}{Common noun} & \multirow{ 2}{*}{Subject} & \multirow{ 2}{*}{Object} & \(p(\text{correct}\mid\text{SO}, \text{m.\hspace{0.1cm}cnoun\hspace{0.1cm}psubj})\) & 28.17 & \multirow{ 2}{*}{95\%} \\
        & & & \(p(\text{correct}\mid\text{SO}, \text{f.\hspace{0.1cm}cnoun\hspace{0.1cm}psubj})\) & 26.22 \\
        \hline
        \multirow{ 2}{*}{Common noun} & \multirow{ 2}{*}{Object} & \multirow{ 2}{*}{Subject} & \(p(\text{correct}\mid\text{SO}, \text{m.\hspace{0.1cm}cnoun\hspace{0.1cm}pobj})\) & 33.11 & \multirow{ 2}{*}{99\%} \\
        & & & \(p(\text{correct}\mid\text{SO}, \text{f.\hspace{0.1cm}cnoun\hspace{0.1cm}pobj})\) & 22.42 \\
        \hline
        \multirow{ 2}{*}{Proper noun} & \multirow{ 2}{*}{Subject} & \multirow{ 2}{*}{Object} & \(p(\text{correct}\mid\text{SO}, \text{m.\hspace{0.1cm}pnoun\hspace{0.1cm}psubj})\) & 27.50 & \multirow{ 2}{*}{n.s.} \\
        & & & \(p(\text{correct}\mid\text{SO}, \text{f.\hspace{0.1cm}pnoun\hspace{0.1cm}psubj})\) & 27.98 \\
        \hline
        \multirow{ 2}{*}{Proper noun} & \multirow{ 2}{*}{Object} & \multirow{ 2}{*}{Subject} & \(p(\text{correct}\mid\text{SO}, \text{m.\hspace{0.1cm}pnoun\hspace{0.1cm}pobj})\) & 23.25 & \multirow{ 2}{*}{n.s.} \\
        & & & \(p(\text{correct}\mid\text{SO}, \text{f.\hspace{0.1cm}pnoun\hspace{0.1cm}pobj})\) & 21.06 \\
        \hline
    \end{tabular}
    \caption[Conditional probabilities for the correctness of predictions according to properties of subjects and objects]{Conditional probabilities for the correctness of predictions given the subject's or the object's gender. The rightmost column indicates a significant difference between compared proportions at the 99\% or 95\% confidence level, or no significance (n.s.), using a z-test for the equality of two proportions. }
    \label{tab:gender-probabilities}
\end{table*} 

\section{Generalization sets}

Tables \ref{tab:wogli-generalization-examples-pronoun}, \ref{tab:wogli-generalization-examples-dative} and \ref{tab:wogli-generalization-examples-ditransitive} provide examples for pairs created for generalization sets.

\begin{table*}[!htb]
\centering
 \begin{tabular}{lllll}
  & \multicolumn{4}{l}{} \\
 \hline
 Premise & \underline{\textbf{Er}}$_{NOM-SG-M}$ & warnt$_{SG}$ & \underline{den}$_{ACC-SG-M}$ & Gast \\
 & \textbf{He}$_{NOM-SG-M}$ & warns$_{SG}$ & the$_{ACC-SG-M}$ & guest \\
 & \textit{He} & \textit{warns} & \textit{the} & \textit{guest} \\
 \hline
 H1-SO (NE) & \underline{\textbf{Der}}$_{NOM-SG-M}$ & \textbf{Gast} &
 warnt$_{SG}$ & \underline{ihn}$_{ACC-SG-M}$ \\
  & \textbf{The}$_{NOM-SG-M}$ & \textbf{guest} & warns$_{SG}$ & him$_{ACC-SG-M}$ \\
  & \textit{The} & \textit{guest} & \textit{warns} & \textit{him} \\
  \hline
  H2-OS (E)* & \underline{Den}$_{ACC-SG-M}$ & Gast & warnt$_{SG}$ & \underline{\textbf{er}}$_{NOM-SG-M}$ \\
  & The$_{ACC-SG-M}$ & guest & warns$_{SG}$ & \textbf{he}$_{NOM-SG-M}$ \\
  & \textit{He} & \textit{warns} & \textit{the} & \textit{guest} \\
 \hline
 \end{tabular}
\caption[Examples of WOGLI-p-subject pairs]{Examples of WOGLI-p-subject pairs. Just as in WOGLI, the entailed hypothesis has a marked word order.}
\label{tab:wogli-generalization-examples-pronoun}
\end{table*}

\begin{table*}[!htb]
 \resizebox{\textwidth}{!}{
 \centering
 \begin{tabular}{llllll}
  & \multicolumn{5}{l}{} \\
 \hline
 Premise & \underline{\textbf{Ein}}$_{NOM-SG-M}$ & \textbf{Richter} & \underline{gratuliert}$_{SG}$ & \underline{diesen}$_{DAT-PL-M}$ & \underline{Beratern}\\
 & \textbf{A}$_{NOM-SG-M}$ & \textbf{judge} & congratulates$_{SG}$ & these$_{DAT-PL-M}$ & consultants \\
 & \textit{A} & \textit{judge} & \textit{congratulates} & \textit{these} & \textit{consultants} \\
 \hline
 H1-SO (NE) & \underline{\textbf{Diese}}$_{NOM-PL-M}$ & \underline{\textbf{Berater}} & \underline{gratulieren}$_{PL}$ & \underline{einem}$_{DAT-SG-M}$ & Richter \\
 & \textbf{These}$_{NOM-PL-M}$ & \textbf{consultants} & congratulate$_{PL}$ & a$_{DAT-SG-M}$ & judge \\
 & \textit{The} & \textit{consultants} & \textit{congratulate} & \textit{a} & \textit{judge} \\
 \hline
 H2-OS (E)* & \underline{Diesen}$_{DAT-PL-M}$ & \underline{Beratern} & \underline{gratuliert}$_{SG}$ & \underline{\textbf{ein}}$_{NOM-SG-M}$ & \textbf{Richter} \\
 & These$_{DAT-PL-M}$ & consultants & congratulates$_{SG}$ & \textbf{a}$_{NOM-SG-M}$ & \textbf{judge}  \\
 & \textit{A} & \textit{judge} & \textit{congratulates} & \textit{these} & \textit{consultants} \\
 \hline
\end{tabular}}
\caption[Examples of WOGLI-dative pairs]{Examples of WOGLI-dative pairs. Just as in WOGLI, the entailed hypothesis has a marked word order.}
\label{tab:wogli-generalization-examples-dative}
\end{table*}

\begin{table*}[!htb]
 \resizebox{\textwidth}{!}{
 \centering
 \begin{tabular}{p{2.5cm}lllllll}
  & \multicolumn{7}{l}{} \\
 \hline
  Premise & \underline{\textbf{Die}}$_{NOM-PL-F}$ & \textbf{Kellnerinnen} & \underline{geben}$_{PL}$ & \underline{einem}$_{DAT-SG-M}$ & Händler & den & Kuchen \\
   & \textbf{The}$_{NOM-PL-F}$ & \textbf{waitresses} & give$_{PL}$ & a$_{DAT-SG-M}$ & merchant & the & cake\\
   & \textit{The} & \textit{waitresses} & \textit{give} & \textit{the} & \textit{cake} & \textit{to a} & \textit{merchant}\\
  \hline
  H1-SiO (NE) & \underline{\textbf{Ein}}$_{NOM-SG-M}$ & \textbf{Händler} & \underline{gibt}$_{SG}$ & \underline{den}$_{DAT-PL-SG}$ & Kellnerinnen & den & Kuchen \\
   & \textbf{A}$_{NOM-SG-M}$ & \textbf{merchant} & gives$_{SG}$ & the$_{DAT-PL-SG}$ & waitresses & the & cake \\
   & \textit{A} & \textit{merchant} & \textit{gives} & \textit{the} & \textit{cake} & \textit{to the} & \textit{waitresses}\\
  \hline
  H2-iOS (E)* & \underline{Einem}$_{DAT-SG-M}$ & Händler & \underline{geben}$_{PL}$ & \underline{\textbf{die}}$_{NOM-PL-F}$ & \textbf{Kellnerinnen} & den & Kuchen \\
   & A$_{DAT-SG-M}$ & merchant & give$_{PL}$ & \textbf{the}$_{NOM-PL-F}$ & \textbf{waitresses} & the & cake \\
   & \textit{The} & \textit{waitresses} & \textit{give} & \textit{the} & \textit{cake} & \textit{to a} & \textit{merchant}\\
 \hline
\end{tabular}}
\caption{Examples of WOGLI-ditransitive pairs. Just as in WOGLI, the entailed hypothesis has a marked word order. 
}
\label{tab:wogli-generalization-examples-ditransitive}
\end{table*}



\end{document}